%% file: HIMO_NIPSOPTRL_CR.tex
\documentclass{article}

\PassOptionsToPackage{numbers, compress}{natbib}

     \usepackage[final]{optrl_2019}

\usepackage[utf8]{inputenc} %
\usepackage[T1]{fontenc}    %
\usepackage{hyperref}       %
\usepackage{url}            %
\usepackage{booktabs}       %
\usepackage{amsfonts}       %
\usepackage{nicefrac}       %
\usepackage{microtype}      %

\usepackage[]{algorithm2e}
\usepackage{amsmath, amssymb}
\usepackage{graphicx}
\usepackage{placeins}
\usepackage{wrapfig}
\usepackage{tabularx}
\usepackage{xcolor}

\usepackage{etoc}

\usepackage[T1]{fontenc}
\usepackage[ttdefault=true]{AnonymousPro}

\usepackage{amsthm}
\theoremstyle{definition}

\newtheorem{theorem}{Theorem}[section]
\newtheorem{proposition}[theorem]{Proposition}

\usepackage{cleveref}
\crefname{theorem}{Theorem}{Theorems}
\crefname{proposition}{Proposition}{Propositions}
\crefname{lemma}{Lemma}{Lemmas}
\crefname{corollary}{Corollary}{Corollarys}

\newcommand{\beginsupplement}{%
        \setcounter{table}{0}
        \renewcommand{\thetable}{S\arabic{table}}%
        \setcounter{figure}{0}
        \renewcommand{\thefigure}{S\arabic{figure}}%
     }
\def\nn{\nonumber}
\def\bsymb{\boldsymbol}
\def\cA{{\cal A}}
\def\cC{{\cal C}}

\def\cI{{\cal I}}
\def\cL{{\cal L}}

\def\cS{{\cal S}}

\def\cX{{\cal X}}

\def\ba{{\mathbf a}}

\def\bn{{\mathbf n}}

\def\bp{{\mathbf p}}
\def\bq{{\mathbf q}}
\def\bs{{\mathbf s}}

\def\bpi{{\bsymb \pi}}

\def\btau{{\boldsymbol \tau}}

\def\R{\mathbb{R}}

\def\bcA{{\boldsymbol \cA}}

\def\bcS{{\boldsymbol \cS}}

\def\cTau{\mathcal{T}}
\def\bcTau{{\bsymb \cTau}}

\def\inv{^{-1}}

\def\inv{^{-1}}

\def\zero{\texttt{0}}

\def\p{\mathrm{p}}

\newcommand{\HH}{\mathrm{H}}

\newcommand{\RR}{\mathrm{R}}

\DeclareMathOperator*{\argmax}{argmax}

\DeclareMathOperator*{\KL}{KL}
\DeclareMathOperator*{\PD}{PD}
\DeclareMathOperator*{\CD}{CD}

\def\ERTrans{R}  %
\def\ERPath{\mathbf{\ERTrans}}  %
\def\ERPathPol{\mathcal{\ERTrans}} %
\def\ap{A} %
\def\cp{C} %
\def\bap{\mathbf{\ap}}

\def\state{s}
\def\action{a}
\def\saction{\tau}
\def\bstate{\bs}
\def\baction{\ba}
\def\bpath{\btau}
\def\statespace{\cS}
\def\actionspace{\cA}
\def\sactionspace{\cTau}
\def\bstatespace{\bcS}
\def\bactionspace{\bcA}
\def\bpathspace{\bcTau}

\title{Hierarchical model-based policy optimization: from actions to action sequences and back}

\author{
  Daniel McNamee \\
  University College London \\
  \texttt{d.mcnamee@ucl.ac.uk}
}

\begin{document}

\maketitle

\begin{abstract}
We develop a normative framework for hierarchical model-based policy optimization based on applying second-order methods in the space of all possible state-action paths. The resulting \emph{natural path gradient} performs policy updates in a manner which is sensitive to the long-range correlational structure of the induced stationary state-action densities.
We demonstrate that the natural path gradient can be computed exactly given an environment dynamics model and depends on expressions akin to higher-order successor representations. In simulation, we show that the priorization of local policy updates in the resulting policy flow indeed reflects the intuitive state-space hierarchy in several toy problems.
\end{abstract}

\input{HIMO1_main_NIPSOPTRL_CR}

\newpage
\bibliographystyle{unsrt}
\bibliography{library_HIMO} %

\newpage
\vspace{7mm}
\begin{center}
\vspace{7mm}
\noindent\makebox[\textwidth]{\rule{\textwidth}{2.0pt}} \\
\vspace{3mm}
{\bf {\LARGE  Hierarchical model-based policy optimization } \\
\vspace{2mm} {\Large Supplementary Material} }
\vspace{5mm}
\noindent\makebox[\textwidth]{\rule{\textwidth}{1.0pt}}
  {\bf Daniel McNamee} \\
  University College London \\
  \texttt{\small d.mcnamee@ucl.ac.uk}
\end{center}

\addcontentsline{toc}{part}{Supplementary Material}
\localtableofcontents*
\beginsupplement

\include{HIMO2_extsim_NIPSOPTRL_CR}

\include{HIMO3_techder_NIPSOPTRL_CR}

\end{document}

%% file: HIMO1_main_NIPSOPTRL_CR.tex
\vspace {-0.5 em}
\section{Introduction}
\vspace {-0.5 em}
Reinforcement learning algorithms can leverage internal models of environment dynamics to facilitate the development of good control policies \citep{Sutton2018}. Dynamic programming methods iteratively implement one-step, full-width backups in order to propagate value information across a state-space representation and facilitate policy updates \citep{Bellman1954}. Stochastic approximations of this approach underpin a wide range of model-free reinforcement learning algorithms which can be enhanced by the ability to query samples from an environment model internally represented within an agent as in the DYNA architecture \citep{Sutton1990}. State-space search strategies apply heuristic principles to efficiently sample multi-step paths from internal models and have formed a core component of recent state-of-the-art game playing agents \citep{Silver2016}. Model-based policy search algorithms can use paths sampled from a model in order to approximate policy gradients \citep{Deisenroth2011a}. Such methods rely on alternating between simulating paths over various horizons and then using this information to improve the policy either directly or based on bootstrapped value estimates \citep{Sutton2018}. In this study, we introduce a hierarchical model-based policy optimization procedure which normatively improves policies in a manner sensitive to the distribution of all future paths without requiring simulations or cached value functions. In our analysis, the central object of interest is not a state-action pair, as is the standard perspective in reinforcement learning in discrete Markov decision processes (MDPs), but complete state-action sequences or \emph{paths} \cite{Bagnell2003,Kappen2012,Theodorou2013}. We show that the MDP objective can be re-written in terms of a log-likelihood over paths and so gradient ascent in the space of policies over paths integrates information over all possible future paths in expectation on each step. Furthermore, the resulting path gradient and path Hessian at a given policy reflect the induced correlational structure between state-actions across time. As a result of these algorithmic features, and in contrast to dynamic programming techniques \citep{Sutton2018}, this method possesses the normative quality that it generates the unique trajectory through policy space which iteratively maximizes the expected cumulative reward obtained on every step.

In Section~\ref{sec:background}, we summarize the mathematical framework of discrete MDPs in the sum-over-paths formalism, define our notation, and select a policy parametrization. In Section~\ref{sec:second_order}, we derive our hierarchical model-based policy optimization algorithm (HIMO). In Section~\ref{sec:sim_analysis}, we apply the algorithm in two simple problems with hierarchical structure and interrogate the resulting policy optimization dynamics. We conclude with a brief discussion in Section~\ref{sec:discussion}.

\section{The exponential representation of policies over paths}
\label{sec:background}

We consider discounted infinite-horizon Markov decision processes defined by the tuple $(\statespace, \actionspace, P, R, \state_\zero)$ comprised of a state-space $\statespace$, an action-space $\actionspace$, a dynamics tensor $P$, a reward function $R$, and an initial state $\state_\zero$. The action set $\actionspace$ is the union of the sets of actions available at each state $\actionspace=\cup_{\state_i\in\statespace}\actionspace_i$. In each state $\state_i\in\statespace$, the agent selects actions $\action_j\in\actionspace_i$ according to a stochastic policy\footnote{HIMO converges to the optimal deterministic policy asymptotically.} $\pi(\action_j|\state_i)\equiv \pi_{ij}$ resulting in a stochastic state transition governed by the dynamics tensor $P(\state_k|\state_i,\action_j)\equiv p_{ijk}$ and the receipt of a reward $R(\state_i,\action_j,\state_k)\equiv R_{ijk}$. Bold-typed notation, $\bstate\in \bstatespace$,  $\baction\in \bactionspace$, and $\bpath\in\bpathspace$ denotes sequences of states $\state\in\statespace$, actions $\action\in \actionspace$, and state-actions $\saction\equiv(s_\saction,a_\saction)\in\sactionspace$ respectively. A valid (i.e., possible under the policy and environment dynamics) state-action sequence $\bpath:=(\ldots, \bstate_t, \baction_t,\bstate_{t+1},\baction_{t+1},\ldots)$ is referred to as a \emph{path}. The path probability $\bq(\bpath)$ is defined as the joint distribution over states $\bstate$ and actions $\baction$
\begin{eqnarray}
        \bq(\bpath) &:=& \prod_{t=0}^\infty \p(\bstate_{t+1}|\bstate_t,\baction_t) \pi(\baction_t|\bstate_t) = \bp(\bstate_{+1}|\bstate,\baction) \bpi(\baction|\bstate) \label{eqn:path_density}
\end{eqnarray}
where
\begin{eqnarray}
  \bpi(\baction|\bstate) &:=& \prod_{t=0}^{\infty} \pi(\baction_t|\bstate_t)~~~~,~~~~
  \bp(\bstate_{+1}|\bstate,\baction) := \prod_{t=0}^{\infty} \p(\bstate_{t+1}|\bstate_{t},\baction_{t}) ~~. \label{eqn:path_policy}
\end{eqnarray}
The discount parameter $\gamma$ is embedded in the environment dynamics tensor $P$ implying that an episode may end with probability $1-\gamma$ on every time step. We represent the policy in terms of natural parameters $\ap_{ij}$ in an exponential parameterization $\ap_{ij}:=\log{\pi_{ij}}$. These natural parameters are examples of action preferences in reinforcement learning parlance \citep{Sutton2018}. In general, $\ap_{ij}(\theta)$ may be parametrized and our formalism extended to policy gradients in $\theta$-space which we will study in future work. In this manuscript, we consider the action preferences as parameters themselves in order to focus on theory exposition and demonstrations of hierarchical processing emergent within planning considered as policy optimization.

To ensure that the policy probabilities $\pi_{ij}=e^{\ap_{ij}}$ take positive values less than one, the action preferences $\ap_{ij}\in \R^-$ are constrained to take negative real values. To ensure that the set of policy probabilities forms a normalized density at each state, we eliminate a redundant action preference at each state. This is accomplished by defining an arbitrary transition probability at each state in terms of the probabilities of alternative transitions at that state. We index this dependent action preference using an $\omega$ subscript, as in $\ap_{ii_\omega}$, in order to distinguish it from the independent action preferences which will be directly modified during policy optimization. Due to the fact that a policy must be normalized at each state, we have
\begin{eqnarray}
  \pi_{ii_\omega} &=& 1 - \sum_{\action_{i_\omega}\neq \action_j\in\actionspace_i} \pi_{ij} ~~. \label{eqn:pi_norm}
\end{eqnarray}
Under this local policy normalization constraint, the ``fixed'' action preferences are equivalently constrained via the log-sum-exp expression
\begin{eqnarray}
  \ap_{ii_\omega} &=& \log \left(1-\sum_{\action_{i_\omega}\neq \action_j \in \actionspace_i}e^{\ap_{ij}} \right) ~~. \label{eqn:av_norm}
\end{eqnarray}

Given a complete action preference parametrization $\ap:=(\ldots, \ap_{ij},\ldots)_{\state_i\in\statespace, \action_j\in \actionspace_i}$, the path policy $\bpi(\btau)\equiv \bpi(\baction|\bstate)$ and the path density $\bq(\btau)$ have the following log-linear forms
\begin{eqnarray}
  \log{\bpi(\btau)} &=& \ap\cdot n(\bpath) = \sum_{\substack{\state_i,\state_k\in\statespace \\ \action_j \in \actionspace_i}} \ap_{ij} n_{ijk}(\bpath) = \sum_{\substack{\state_i\in\statespace \\ \action_j\in \actionspace_i}} \ap_{ij} n_{ij}(\bpath)  \nn \\
  \log{\bp(\bstate_{+1}|\bstate,\baction)} &=& \cp\cdot n(\bpath) = \sum_{\substack{\state_i,\state_k\in\statespace \\ \action_j \in \actionspace_i}}\cp_{ijk}n_{ijk}(\bpath) \nn \\
  \log{\bq(\bpath)} &=& \ap\cdot n(\bpath) + \cp\cdot n(\bpath) \label{eqn:log_linear}
\end{eqnarray}
where $n_{ijk}$ counts the number of occurrences of the state-action-state event $(\state_i,\action_j,\state_k)$ in the path $\btau$ and $\cp_{ijk}:=\log{p_{ijk}}$. Considering the set of probabilities $e^{(\ap+\cp)\cdot n(\bpath)}$ parametrized by $\ap$ as an exponential family \citep{Nagaoka2005}, the vector $n(\bpath)$ of transition counters $n_{ijk}(\bpath)$ constitutes a sufficient statistic for the path $\bpath$. These counter variables will be crucial in computing gradients with respect to the stationary state-action density induced by a given policy. This, in turn, will facilitate a novel \emph{path gradient} relating policy gradients across distinct states. We now turn to optimizing the policy objective in this policy representation.

\section{Second-order policy optimization in the space of paths}
\label{sec:second_order}

The cumulative expected reward objective as a sum-over-paths \cite{Theodorou2013} is
\begin{eqnarray}
    \ap^* &:=& \argmax_\ap \ERPathPol \left(\ap\right) \nn \\
    \ERPathPol\left(\ap\right) &:=& \left\langle\ERPath(\bpath)\right\rangle_{\bq} ~~,~~\ERPath(\bpath) = \sum_{t=0}^{\infty} \RR(\bstate_{t},\baction_{t}, \bstate_{t+1}) \label{eqn:MDPobj}
\end{eqnarray}
where the angled brackets $\langle\cdot\rangle_{\bq}$ denote the expectation operation over the path density $\bq$ (Eqn.~\ref{eqn:path_density}). Consider the gradient $\nabla_{\ap} \ERPathPol$ of $\ERPathPol$ with respect to the action preferences $\ap$:
\begin{eqnarray}
 \nabla_{\ap} \ERPathPol\left(\ap\right) &=& \sum_{\bpath\in\bpathspace} \nabla_{\ap} \bq(\bpath) \ERPath(\bpath) \nn \\
 &=& \sum_{\bpath\in\bpathspace} \bq(\bpath)\left[ \nabla_{\ap} \log{\bq(\bpath)}\right] \ERPath(\bpath) \nn \\
 &=& \left\langle \nabla_{\ap} \log{\left[\bq(\bpath)e^{\ERPath(\bpath)}\right]}  \right\rangle_{\bq} ~~. \label{eqn:like_grad}
\end{eqnarray}
It is observed that the gradient $\nabla_{\ap} \ERPathPol\left(\ap\right)$ has the form of a score function in expectation over paths with $\cL_{\bpath}(\ap):=\bq(\bpath)e^{\ERPath(\bpath)}$ playing the role of a path likelihood function.

Our theoretic hypothesis is that the curvature information utilized in a second-order optimization procedure \cite{Boyd2004a} of the path-based objective (Eqn.~\ref{eqn:MDPobj}) will enforce policy improvements which are sensitive to the state-action correlational structure over all horizons embedded in the path density $\bq(\bpath)$. Therefore, we implement the Newton step
\begin{eqnarray}
    \ap^{t+1} &\leftarrow & \ap^{t} + \left\langle \HH_{\cL_\bpath}\left(\ap^t\right) \right\rangle_{\bq}\inv \nabla_{\ap}\ERPathPol\left(\ap^t\right) ~~. \label{eqn:exp_update}
\end{eqnarray}
where $\left\langle \HH_{\cL_\bpath}\left(\ap^t\right) \right\rangle_{\bq}$ is the expected Hessian of the path likelihood function $\cL_{\bpath}(\ap)$ at the current action preferences $\ap^t$. As a trust region method \cite{Boyd2004a}, this can be interpreted as optimizing a parameter step $\Delta \ap$ under a local approximation to $\ERPathPol\left(\ap^t\right)$:
\begin{eqnarray}
    \Delta \ap^* &=& \argmax_{\Delta \ap} \left[\nabla_\ap \ERPathPol\left(\ap^t\right)\cdot \Delta \ap + \frac{1}{2}||\Delta \ap||^2_{\left\langle \HH_{\cL_\bpath}\left(\ap^t\right) \right\rangle_{\bq}}\right] \label{eqn:second_order}
\end{eqnarray}
where $||\cdot||_{\left\langle \HH_{\cL_\bpath}\left(\ap^t\right) \right\rangle_{\bq}}$ is the expected Hessian norm of the path likelihood function $\cL_\bpath$ at $\ap^t$.
It can be shown \cite{Amari1998,Martens2014} that the expectation of the likelihood Hessian (in this case over paths) is equivalent to a Fisher information expression $\left\langle \HH_{\cL_\bpath}\left(\ap\right) \right\rangle_{\bq}\equiv \cI(\ap)$ with components\footnote{The Fisher information matrix $\cI$ for the path likelihood is derived in Section~\ref{sec:exp_param_fisherinfo} of the Supplementary Material (SM).}
\begin{eqnarray}
  \left[\cI(\ap)\right]_{ij,kl} &:=&  \left\langle \left[\partial_{\ap_{ij}} \log{\bq(\bpath)} \right]\left[\partial_{\ap_{kl}} \log{\bq(\bpath)} \right] \right\rangle_{\bq} ~~.
\end{eqnarray}
Therefore, the \emph{natural path gradient} step
\begin{eqnarray}
    \ap^{t+1} &\leftarrow & \ap^{t} + \cI\inv\left(\ap^t\right) \nabla_{\ap}\ERPathPol\left(\ap^t\right) \label{eqn:exp_update_fi}
\end{eqnarray}
is equivalent to the Newton step (Eqn.~\ref{eqn:exp_update}) and is optimal with respect to the local path-based approximation of the cumulative reward objective (Eqn.~\ref{eqn:second_order}). Natural gradients implement parametrization independent, or covariant, updates \cite{Amari1998}. Therefore, our planning-as-optimization scheme is equivalent to natural gradient ascent in the path probability $\bpi$ parametrization of the cumulative reward objective $\ERPathPol$. Since $\ERPathPol$ is convex in $\bpi$, natural path gradient ascent (Eqn.~\ref{eqn:exp_update_fi}) can be expected to converge to the globally optimal policy and does so in all simulated problems. In the SM (Section~\ref{sec:tech_derivations}), we show that exact expressions for all the quantities required to implement this policy optimization procedure can be derived analytically. Notably, the policy path gradient is a function of state-action correlation functions which are themselves composed of products of successor representations \cite{Dayan1993}. Based on these derivations, the complete algorithm which implements the hierarchical model-based path gradient updates (Eqn.~\ref{eqn:exp_update}) is elaborated in subsection~\ref{sec:algo} (SM).

\section{Simulations}
\label{sec:sim_analysis}

We simulate hierarchical model-based policy optimization (Eqn.~\ref{eqn:path_gradient}) in two toy environments in order to gain insight into the dynamics of the policy optimization process and demonstrate its hierarchical sensibility. Specifically, the Tower of Hanoi problem and a four-room grid world environment.
After running HIMO until convergence, its dynamics are interrogated using two measures. The first measure is the KL-divergence between the policy densities at each iteration $\pi^t$ and the prior policy $\pi^0$. We compute this \emph{policy divergence} measure $\PD$ locally at each state $\state \in \statespace$:
\begin{equation}
\PD(\state,t) := \KL\left[\pi^t_{ \state \cdot } ||\pi^0_{\state \cdot} \right] ~~. \label{eqn:policy_div}
\end{equation}
Policy divergence quantifies the degree to which the algorithm is modifying the local policy at each state as a function of planning time. The second measure is the difference between the expected number of times a state will be occupied under the currently optimized policy versus the prior policy. Specifically, the \emph{counter difference} measure $\CD$ is
\begin{equation}
\CD(\state,t) := D_{\zero \state}^t-D_{\zero \state}^0 ~~.  \label{eqn:counter_diff}
\end{equation}
where $D^t$ is the successor representation under policy $\pi^t$ \cite{Dayan1993}. Counter differences indicate how HIMO prioritizes visits to, or avoidance of, states as a function of planning time. We study these measures as well as their time derivatives in their original form as well as after max-normalizing per state in order to facilitate comparisons across states:
\begin{eqnarray}
\widetilde{\PD}(\state,t) := \frac{\PD(\state,t)}{\max_t\PD(\state,t)} ~~,~~ \widetilde{\CD}(\state,t) :=\frac{\CD(\state,t)}{\max_t|\CD(\state,t)|} ~~.
\end{eqnarray}
The Tower of Hanoi example highlights the intuitive hierarchical qualities of the algorithm, and, in the room-world, the capacity of the algorithm to radically alter its dynamics in response to the relatively minor modifications of the state-space is observed.

\begin{figure}[ht!]
\centering
\includegraphics[width=\textwidth]{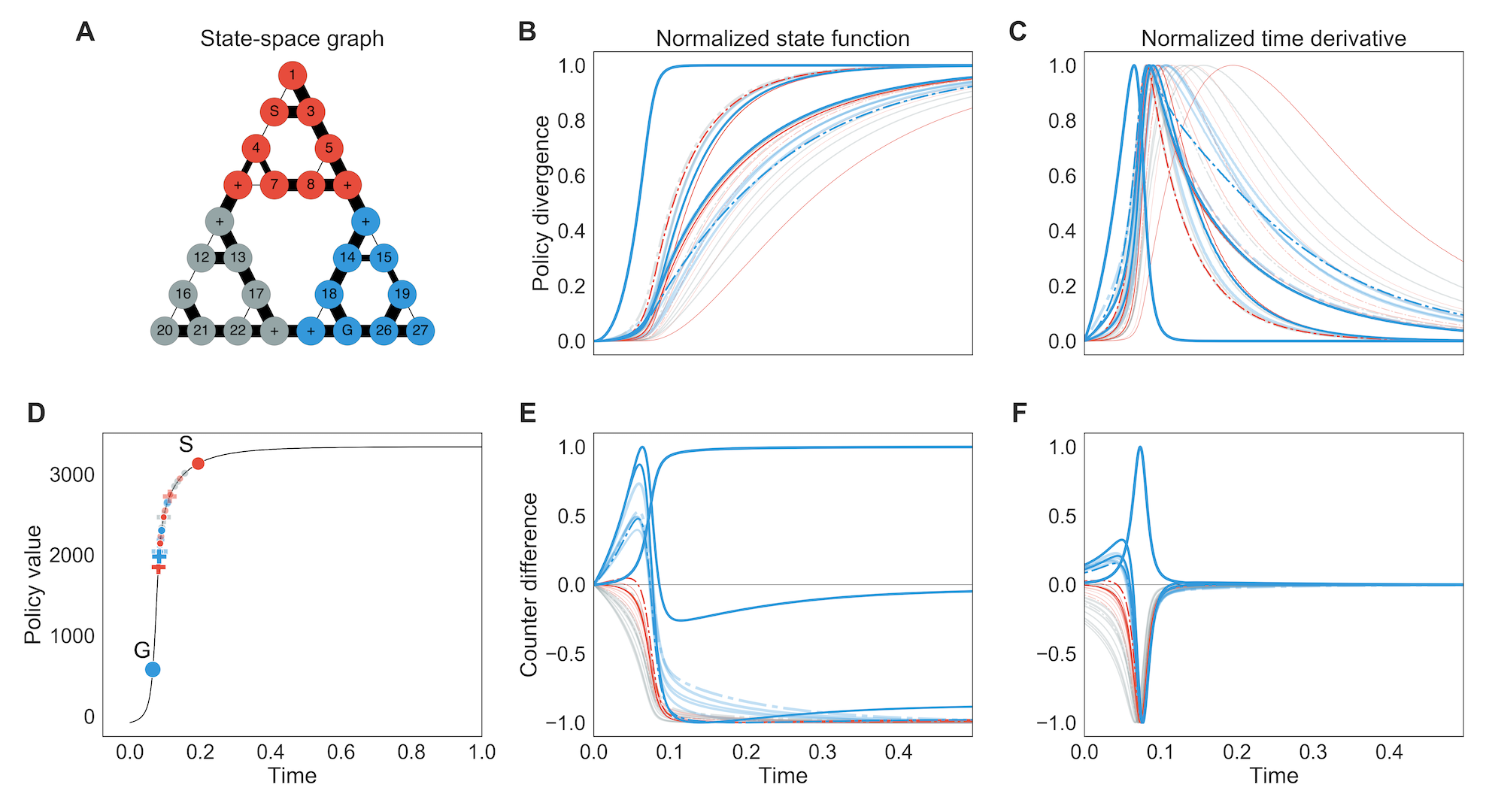}
\caption{\textbf{Hierarchical policy optimization dynamics in the Tower of Hanoi game.} \textbf{A.} Tower of Hanoi state-space graph. The problem is to find the shortest path from the start state $S$ to the goal state $G$. \textbf{B-C.} Normalized policy divergence $\widetilde{\PD}$ and its time derivative for each state. The color of the curve indicates which state it corresponds to in panel A. Dotted lines correspond to bottleneck states marked + in panel A. Lines for states which are not along the optimal path are plotted with partial transparency. \textbf{D.} Policy value as a function of planning time. Time-to-max policy divergence velocities (i.e. the peaks of the curves in panel C) are dotted along the policy value curve for states along the optimal path. \textbf{E-F.} Normalized counter difference $\widetilde{\CD}$ and its time derivative.}
\label{fig:TOH_compact}
\end{figure}

In our Tower of Hanoi simulation (Fig.~\ref{fig:TOH_compact}), the agent is endowed with the ability to remain at a state thus the optimal policy is to transit to state G and then choose to remain there (since it can then accumulate further reward on every time step). This choice to remain is interpreted as a ``STOP'' signal in that the agent has decided to terminate its environment interactions. Of all actions in all states in the environment, HIMO policy improvements prioritize this action to ``STOP'' at the goal state. This can be observed in the relatively rapid policy divergence $\widetilde{\PD}$ at the goal state (Fig.~\ref{fig:TOH_compact}B) and the fact that the policy divergence velocity peaks for the goal state before all others (Fig.~\ref{fig:TOH_compact}D). Focusing on states along the optimal path only, the second highest priority is assigned to the bottleneck between the red and blue clusters. Local policy optimization at the start state is deferred to last. Through the counter difference measure $\widetilde{\CD}$, we can observe how HIMO increases the occupation rate of all states in the same cluster as the goal state (in blue) before subsequently avoiding non-goal states in the goal cluster (Fig.~\ref{fig:TOH_compact}E). These non-monotonic counter difference trajectories suggest that HIMO treats all blue states as a unitary abstraction initially before refining its planning strategy to distinguish individual states within the goal cluster. The increasing spatial resolution at which the algorithm distinguishes states over time, and the priorization pattern of local policy adaptations, show how HIMO is dynamically sensitive to the hierarchical structure of the state-space.

\begin{figure}[h!]
\centering
\includegraphics[width=\textwidth]{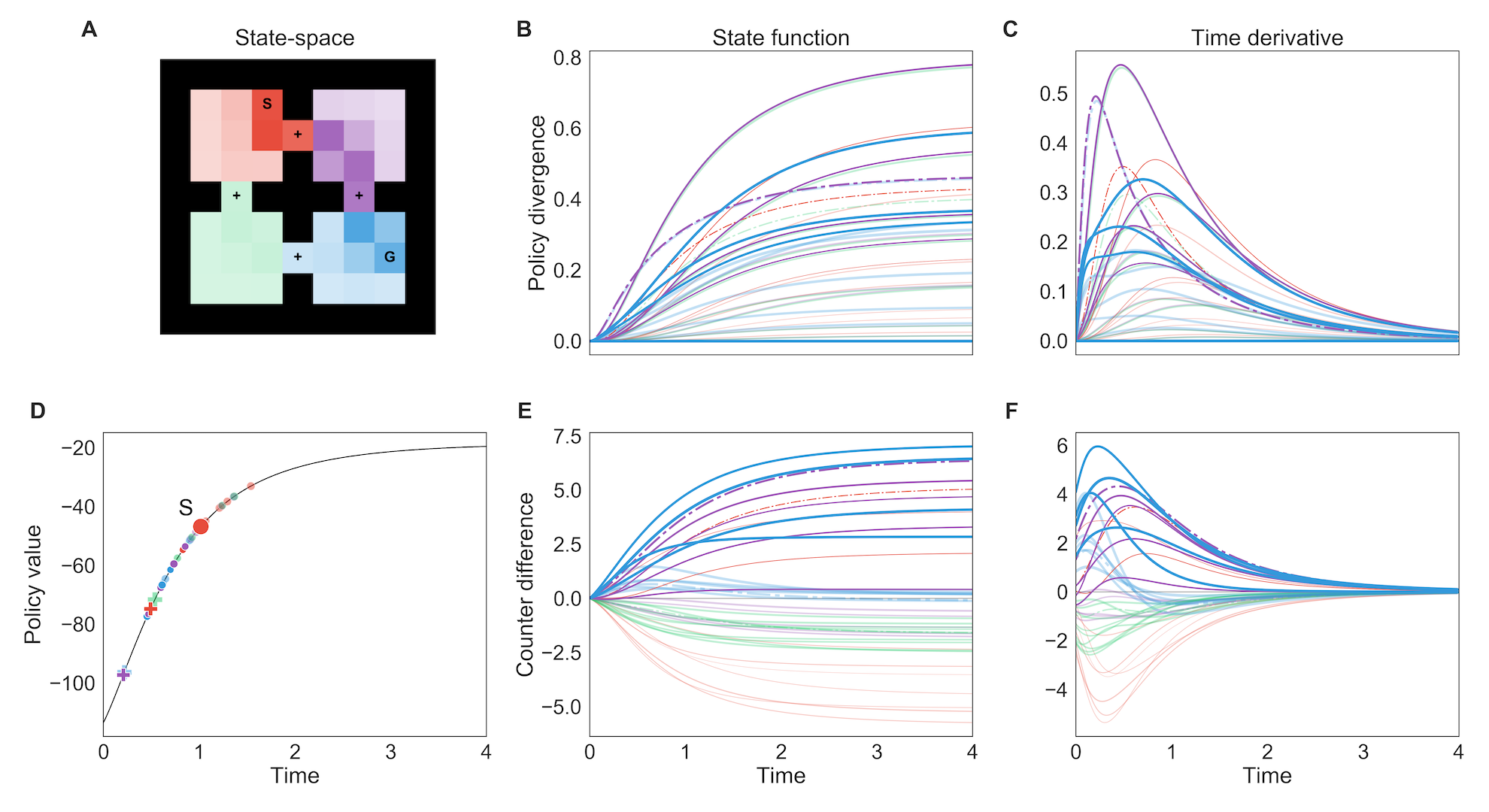}
\caption{\textbf{Model-based policy optimization the optimal policy in a grid world.} Panels as in Fig.~\ref{fig:TOH_compact} but without policy divergence $\PD$ and counter difference $\CD$ normalization. Darker state colors indicate higher densities of state occupation under the optimal policy.}
\label{fig:room_world}
\end{figure}

In the room world simulation (Fig.~\ref{fig:room_world}), the agent must navigate from the start state S in the northwest room to the goal state G in the southeast room (Fig.~\ref{fig:room_world}A). HIMO prioritizes the adaptation of the local policies at the critical bottleneck states entering the goal room (Fig.~\ref{fig:room_world}D). In contrast, when a ``wormhole'' is available (Fig.~\ref{fig:room_world_wormhole}, SM), the algorithm avoids the, now suboptimal, route through the doorways and prioritizes the local adaptation of the policies at the entrance and exit of the wormhole.

\FloatBarrier

\FloatBarrier

\section{Discussion}
\label{sec:discussion}

We introduced a novel model-based policy optimization procedure which we demonstrated to be sensitive to the hierarchical structure of the state-action space of the MDP. This feature is due to the use of second-order gradient information drawn from a sum-over-paths representation of the MDP objective. Natural path gradients improve policies along the steepest ascent trajectory where the policy metric is induced from the space of policies defined over state-action paths.

In previous work, natural policy gradient and actor-critic methods \cite{Kakade2001, Bagnell2003, Peters2005} have modified standard policy gradient steps using Fisher information matrices in order to perform parametrized policy updates in a manner that is sensitive to the KL-divergence between old and new local policies on average at each state. However, by averaging distinct natural gradients localized at each state weighted by the stationary state distribution, these methods do not relate the components of the policy gradient across states or state-actions as in HIMO and thus are agnostic to the hierarchical structure of the state-space. As previously mentioned, the action preferences $\ap$ employed in HIMO may be parametrized e.g. $\ap(\theta)$. Therefore, the natural path gradient in terms of $\ap$ (sensitive to the global structure across states) established here may be combined with the natural gradient in terms of $\theta$ with respect to state-action probabilities (reflecting local structure at each state) through the reparametrization rule for Fisher informations.

Hierarchical reinforcement learning refers to the acquisition and use of hierarchical state or policy representations which can aid RL agents in overcoming the curse of dimensionality amongst other benefits \cite{Barto2003}. Theoretic approaches to defining optimal hierarchies typically make use of ad hoc objectives which do not pertain directly to the fundamental goal of policy improvement \citep{VanDijk2011b,Solway2014,McNamee2016}. This is in contrast to the intuitive hierarchies emergent in the normative dynamics of HIMO. This motivates its use as a theoretic tool for analyzing the hierarchical structure of policy space since functional relationships between actions over all spatiotemporal scales are explicitly embedded within policy path gradients. In the classic hierarchical tasks simulated here, HIMO dynamically clusters then distinguishes state occupation densities (Fig.~\ref{fig:TOH_compact}), implicitly prioritizes policy improvements at critical bottleneck states (Fig.~\ref{fig:room_world}), and restructures the policy flow in order to take advantage of shortcuts when available at the earliest stages of processing (Fig.~\ref{fig:room_world_wormhole}). All such effects emerge from the single normative principle of performing policy gradient ascent in path space (Eqn.~\ref{eqn:exp_update}). Whereas these effects are manifest in the output of the algorithm, it will be informative to explore its internal dynamics. For example, characterizing the evolution of the counter correlation functions and expected Hessians as a function of planning time may provide insights into how scalable approximations to HIMO may be implemented.

%% file: HIMO2_extsim_NIPSOPTRL_CR.tex
\newpage
\section{Extended simulations}

\subsection{Room world with wormhole}

\begin{figure}[ht!]
\centering
\includegraphics[width=\textwidth]{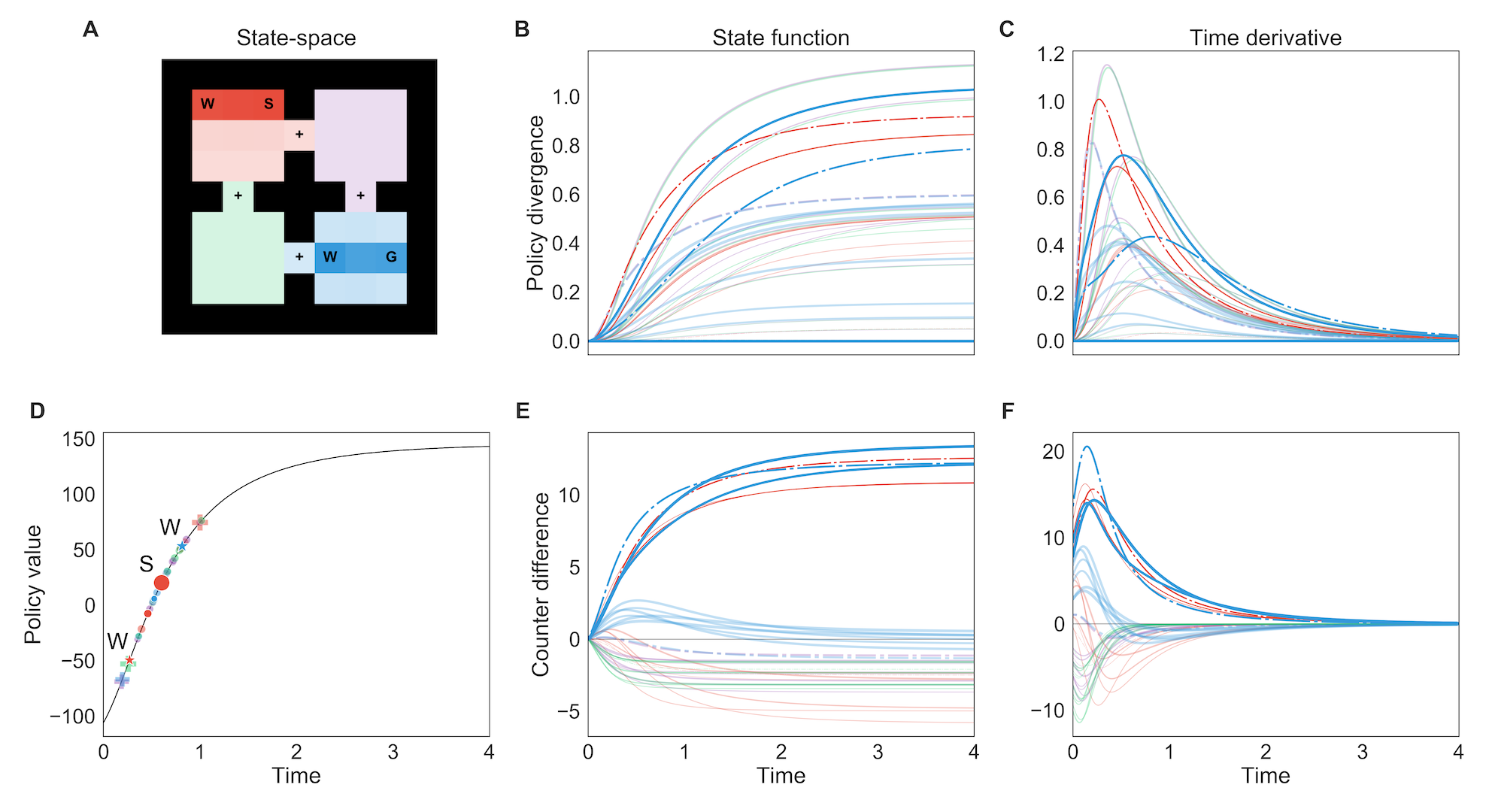}
\caption{\textbf{Hierarchical model-based policy optimization in a grid world with a wormhole.} Panels as in Fig.~\ref{fig:room_world}. Dotted lines with short dashes correspond to bottleneck states marked + in panel \textbf{A}. Dotted lines with long dashes correspond to wormhole states marked W in panel A. Darker state colors indicate higher densities of state occupation under the optimal policy.}
\label{fig:room_world_wormhole}
\end{figure}

%% file: HIMO3_techder_NIPSOPTRL_CR.tex
\newpage
\section{Technical derivations}
\label{sec:tech_derivations}

\subsection{Partial derivatives}
\label{sec:exp_param_pathgrad}

In this section, we record some complementary calculations (Eqn.~\ref{eqn:exp_param_pathgrad_fixed}) and a proposition (\ref{prop:exp_param_policy_partial}) which will be applied in Theorem ~\ref{thm:exp_param_pathgrad}. The partial derivative $\partial_{\ap_{ij}}\ap_{kk_\omega}$ of a dependent action preference $\ap_{kk_\omega}$ with respect to an independent action preference $\ap_{ij}$ is
\begin{eqnarray}
  \partial_{\ap_{ij}}\ap_{kk_\omega} &=& \partial_{\ap_{ij}} \left[\log \left(1-\sum_{\action_{k_\omega}\neq \action_l \in \actionspace_k}e^{\ap_{kl}} \right)  \right] \nn \\
  &=& \left(1-\sum_{\action_{k_\omega}\neq \action_l \in \actionspace_k}e^{\ap_{kl}} \right)\inv \delta_{ik} \left[-e^{\ap_{kj}} \right] \nn \\
  &=& -\delta_{ik}e^{\ap_{kj}-\ap_{kk_\omega}} ~~. \label{eqn:exp_param_pathgrad_fixed}
\end{eqnarray}

\begin{proposition}
  The partial derivative of the path density $\bq(\bpath)$ and the path policy $\bpi(\bpath)$ with respect to $\ap_{ij}$ is
\begin{eqnarray}
  \partial_{\ap_{ij}} \bq(\bpath) &=& \bq(\bpath) \partial_{\ap_{ij}} \log{\bq(\bpath)} \nn \\
  &=&  \bq(\bpath) \partial_{\ap_{ij}} \log{\bpi(\bpath)} \nn \\
  &=& \bq(\bpath)\left[n_{ij}(\bpath) - e^{\ap_{ij}-\ap_{ii_\omega}} n_{ii_\omega}(\bpath) \right] ~~. \label{eqn:exp_param_policy_partial}
\end{eqnarray}
\label{prop:exp_param_policy_partial}
\end{proposition}
\begin{proof}
Using the log-derivative trick
\begin{eqnarray}
  \partial_{\ap_{ij}}\bq(\bpath) &=& \bq(\bpath)\partial_{\ap_{ij}}\left[\log{\bq(\bpath)}\right] \nn \\
   &=& \bq(\bpath)\partial_{\ap_{ij}}\left[\log{\bpi(\bpath)}+\log{\bp(\bstate_{+1}|\bstate,\baction)}\right] \nn \\
  &=& \bq(\bpath)\partial_{\ap_{ij}}\left[\bap\cdot\bn(\bpath) \right] \nn \\
  &=& \bq(\bpath)\partial_{\ap_{ij}}\left\{\sum_{k\in\cX} \left[\sum_{k_\omega\neq l\in \cX_k}\ap_{kl}n_{kl}(\bpath) + \ap_{kk_\omega}n_{kk_\omega}(\bpath)\right]\right\} \nn \\
  &=& \bq(\bpath)\left[n_{ij}(\bpath) + \left(\partial_{\ap_{ij}}\ap_{ii_\omega}\right) n_{ii_\omega}(\bpath) \right] \nn \\
  &=& \bq(\bpath)\left[n_{ij}(\bpath) - e^{\ap_{ij}-\ap_{ii_\omega}}  n_{ii_\omega}(\bpath) \right]
\end{eqnarray}
re-using Eqn.~\ref{eqn:exp_param_pathgrad_fixed}.
\end{proof}

\subsection{Model-based policy path gradient}
\label{sec:path_grad_obj}

\begin{theorem}
  The policy path gradient in the exponential parametrization is defined by the partial derivatives
  \begin{eqnarray}
    \partial_{\ap_{ij}} \ERPathPol\left(\ap\right) &=& \sum_{\substack{\state_k\in\statespace \\ \action_l\in \actionspace_k}}\left[\cC_{ij,kl}- e^{\ap_{ij}-\ap_{ii_\omega}} \cC_{ii_\omega,kl} \right]\RR_{kl}
  \end{eqnarray}
  where $\cC_{ij,kl}:=\left\langle n_{ij}(\bpath) n_{kl}(\bpath)\right\rangle_{\bq}$ are state-action counter correlations and $\RR_{kl}:=\left\langle \RR(\state_i,\action_j,\state_k) \right\rangle_{\p(\state_k|\state_i,\action_j)}$.
  \label{thm:exp_param_pathgrad}
\end{theorem}
\begin{proof}
\begin{align}
  \partial_{\ap_{ij}} \ERPathPol\left(\ap\right) &= \partial_{\ap_{ij}} \sum_{\bpath\in\bpathspace} \bq(\bpath)\left[ \nabla_{\ap} \log{\bq(\bpath)}\right] \ERPath(\bpath) \nn \\
  &= \sum_{\bpath\in\bpathspace} \bq(\bpath)\left[n_{ij}(\bpath) - e^{\ap_{ij}-\ap_{ii_\omega}} n_{ii_\omega}(\bpath)  \right] \ERPath(\bpath) \tag{$\Leftarrow$ Proposition~\ref{prop:exp_param_policy_partial}} \\
  &= \sum_{\bpath\in\bpathspace} \left\{\bq(\bpath)\left[n_{ij}(\bpath) - e^{\ap_{ij}-\ap_{ii_\omega}} n_{ii_\omega}(\bpath)  \right]\right\} \left[\sum_{\substack{\state_k,\state_m \in\statespace \\ \action_l\in \actionspace_k}}n_{klm}(\bpath)\ERTrans_{klm} \right] \notag \\
  &= \sum_{\substack{\state_k,\state_m \in\statespace \\ \action_l\in \actionspace_k}}\left[\left\langle n_{ij}(\bpath)n_{klm}(\bpath)\right\rangle_{\bq}-e^{\ap_{ij}-\ap_{ii_\omega}} \left\langle n_{ii_\omega}(\bpath)n_{klm}(\bpath)\right\rangle_{\bq} \right]\ERTrans_{klm} \notag \\
  &= \sum_{\substack{\state_k,\state_m\in\statespace \\ \action_l\in \actionspace_k}}\left[\cC_{ij,kl}- e^{\ap_{ij}-\ap_{ii_\omega}} \cC_{ii_\omega,kl} \right]p_{klm }\ERTrans_{klm} \notag \\
  &= \sum_{\substack{\state_k\in\statespace \\ \action_l\in \actionspace_k}}\left[\cC_{ij,kl}- e^{\ap_{ij}-\ap_{ii_\omega}} \cC_{ii_\omega,kl} \right]\ERTrans_{kl}  ~~. \notag
\end{align}
\end{proof}
The path gradient depends on the state-action \emph{counter correlations} $\cC_{ij,kl}:=\left\langle n_{ij}(\bpath) n_{kl}(\bpath)\right\rangle_{\bq}$. Given a policy $\pi$ and an environment dynamics model $P$, the state-action correlation functions $\cC$ can be derived using Markov chain theory \cite{Kemeny1983} and therefore the policy path gradient can be expressed analytically. In order to ensure convergence of these quantities, we assume a discounted horizon with \emph{foresight} parameter $\lambda$ which is upper bounded by the discount parameter $\lambda \leq \gamma $. That is, from the perspective of the agent, an episode may end on every timestep with probability $1-\lambda$. This parameter may reflect a limitation on how far the agent can ``see'' into the future.

\subsection{Fisher information}
\label{sec:exp_param_fisherinfo}

State-action selection rates are not independent. Modifying one state-action selection rate under the policy $\pi$ may change the selection rate of another state-action. This is in contrast to the expected reward objective in path space where policy modifications are independent along each path (apart from an overall normalization factor). In order to identify a policy gradient in state-action space with independent gradient components, we will transform the gradient derived in Section~\ref{sec:path_grad_obj} into the natural path gradient pulled back to state-action space.  This is accomplished by pre-multiplying the path gradient by the inverse Fisher information $\cI\inv$ \cite{Amari1998} which, here, relates policy densities in path space $\bpi$ and state-action space $\pi$. This is equivalent to implementing Newton's method as described in the main text. The Fisher information matrix $\cI$ has components
\begin{eqnarray}
  \cI_{ij,kl} &:=& \left\langle \left[\partial_{\ap_{ij}} \log{\bpi(\bpath)} \right]\left[\partial_{\ap_{kl}} \log{\bpi(\bpath)} \right] \right\rangle_{\bq}  \nn \\
  &=& \left\langle \left[n_{ij}(\bpath) - e^{\ap_{ij}-\ap_{ii_\omega}}  n_{ii_\omega}(\bpath) \right]\left[n_{kl}(\bpath) - e^{\ap_{kl}-\ap_{kk_\omega}}  n_{kk_\omega}(\bpath) \right] \right\rangle_{\bq} \nn \\
  &=& \left\langle n_{ij}(\bpath) n_{kl}(\bpath)\right\rangle_{\bpi} -e^{\ap_{kl}-\ap_{kk_\omega}}  \left\langle n_{ij}(\bpath) n_{kk_\omega}(\bpath)\right\rangle_{\bq} + \nn \\
  & & - e^{\ap_{ij}-\ap_{ii_\omega}}  \left\langle n_{kl}(\bpath) n_{ii_\omega}(\bpath)\right\rangle_{\bpi} +  e^{\ap_{ij}-\ap_{ii_\omega}}  e^{\ap_{kl}-\ap_{kk_\omega}}  \left\langle n_{ii_\omega}(\bpath) n_{kk_\omega}(\bpath)\right\rangle_{\bq}  \nn \\
  &=& \cC_{ij,kl} - e^{\ap_{kl}-\ap_{kk_\omega}} \cC_{ij,kk_\omega} - e^{\ap_{ij}-\ap_{ii_\omega}} \cC_{kl,ii_\omega} + e^{\ap_{ij}+\ap_{kl}-\ap_{ii_\omega}-\ap_{kk_\omega}} \cC_{ii_\omega,kk_\omega} ~~. \nn \\
\end{eqnarray}

\newpage
\subsection{Algorithm}
\label{sec:algo}

\begin{algorithm}[H]
  \SetKwInOut{Input}{input}\SetKwInOut{Output}{output}
  \SetAlgoLined
  \textbf{input:} initialized policy $\pi$ with corresponding action preferences $\ap_{ij}:=\log{\pi_{ij}}$ \\
  \While{$\pi$ not converged}{
      \textbf{compute state-state ($D$) and (state-action)-state ($E$) counter correlations \\}
      \ForEach{$(\state_i,\state_k)\in \statespace\times \statespace$}{
      \begin{eqnarray}
       T_{ik}& = & \sum_{a_{j}\in \cA_i}\pi_{ij}p_{ijk} \nn \\
       D_{ik} &=& \left[\left(I- \lambda T\right)\inv\right]_{ik} \nn
      \end{eqnarray}
            \ForEach{$(\action_j)\in \actionspace_i$}{
            \begin{eqnarray}
            E_{(ij)k}&=& \sum_{\state_{k'}\in\statespace} p_{ijk'}D_{k'k} \nn
            \end{eqnarray}
            }
      }
      \textbf{construct (state-action)-(state-action) counter correlations $\cC$ and Fisher information $\cI$ \\}
      \ForEach{$(\state_i,\action_j,\state_k,\action_l)\in \statespace\times \actionspace \times \statespace\times \actionspace$}{
        \begin{eqnarray}
         \cC_{ij,kl} &=& D_{\zero i} \pi_{ij} \delta_{ik} \delta_{jl} + \left[D_{\zero i} E_{(ij)k} + D_{\zero k} E_{(kl)i} \right] \pi_{ij} \pi_{kl} \nn \\
         \cI_{ij,kl} &=& \cC_{ij,kl} - \pi_{kl} \pi_{kk_\omega}\inv \cC_{ij,kk_\omega} - \pi_{ij} \pi_{ii_\omega}\inv \cC_{kl,ii_\omega} + \pi_{ij} \pi_{ii_\omega}\inv \pi_{kl} \pi_{kk_\omega}\inv \cC_{ii_\omega,kk_\omega} \nn
         \end{eqnarray}
         }
         \textbf{update action preferences \\}
        \ForEach{$(\state_i,\action_j)\in \statespace\times \actionspace$}{
         \begin{eqnarray}
        \ap_{ij} &\leftarrow & \ap_{ij} + \sum_{\state_m\in\statespace,\action_n\in\actionspace_m}\left[\cI\inv\right]_{ij,mn} \left\{\sum_{\state_k\in\statespace,\action_l\in \actionspace_k}\left[\cC_{mn,kl}- \pi_{mn} \pi_{mm_\omega}\inv \cC_{mm_\omega,kl} \right]\ERTrans_{kl} \right\} \nn \\ \label{eqn:path_gradient}
        \end{eqnarray}
      }
      \textbf{compute dependent action preferences via normalization \\}
     \ForEach{$\state_i \in \statespace$}{
      \begin{eqnarray}
      \ap_{ii_\omega} &=& \log \left(1-\sum_{\action_{i_\omega}\neq \action_j \in \actionspace_i}e^{\ap_{ij}} \right) \nn
     \end{eqnarray}
     }
      \textbf{express policy \\}
      \ForEach{$(\state_i,\action_j)\in \statespace\times \actionspace$}{
      \begin{eqnarray}
      \pi_{ij} &=& e^{\ap_{ij}} \nn
      \end{eqnarray}
      }
  }
  \Output{optimal policy $\pi^*$}
  \caption{Hierarchical model-based policy optimization in the exponential parametrization. \label{algo:HIMO}}
\end{algorithm}
Note that $(ij)$ is a univariate index of state-action $(\state_i,\action_j)$ combinations. The parameter $\lambda$ is the free parameter $0<\lambda<1$ controlling the agent's foresight.
The path gradient (Eqn.~\ref{eqn:path_gradient}) has several intuitive properties. The matrix $D$ is the successor representation \citep{Dayan1993}. An entry $D_{ij}$ counts the expected number of times that state $\state_j$ will be occupied after starting from state $\state_i$. Therefore the counter correlations $\cC$, which are quadratic in successor representation components, reflect the rate of co-occurrence of pairs of state-actions on average under the policy-generated path distribution. This enables the algorithm to understand the correlative structure of state-action selections under the current policy as proposed. For example, if a temporally remote state-action $(\state_k,\action_l)$ has high reward $\RR_{kl}$ and if there is a high counter correlation $\cC_{ij,kl}$ between a local state-action $(\state_i,\action_j)$ and the remote action (over all horizons), then the reward $\RR_{kl}$ associated with the remote action will be weighted heavily in the path gradient update and added to the local action preference $\ap_{ij}$. The magnitude of this backup is explicitly normalized with respect to a baseline counter correlation $\cC_{ii_\omega,kl}$ associated with the dependent action preference. That is, if the action $(\state_i,\action_{i_\omega})$ is also strongly correlated with $(\state_k,\action_l)$ then the backup to $\ap_{ij}$ is attenuated since the unique contribution of $(\state_i,\action_j)$ in generating $(\state_k,\action_l)$ is diminished. Using such attributional logic, hierarchical model-based policy optimization updates action preferences based on the degree to which a state-action independently generate rewarding paths.

\subsection{Initialization}
\label{sec:exp_param_init}

The prior policy $\pi^0$ can be set to any policy with corresponding initial action preferences
\begin{eqnarray}
  \ap_{ij}^0 &=& \log{\pi_{ij}^0} ~~.
\end{eqnarray}
Assuming that $\pi^0$ is initialized at the random policy, we have
\begin{eqnarray}
  \pi_{ij}^0 &=& \frac{1}{|\actionspace_i|} \nn \\
  \ap_{ij}^0 &=& -\log{|\actionspace_i|}
\end{eqnarray}
for all $\action_j\in\actionspace_i$ for all $\state_i\in\statespace$.